\documentclass[conference]{IEEEtran}
\IEEEoverridecommandlockouts
\usepackage{cite}
\usepackage{amsmath,amssymb,amsfonts}
\usepackage{algorithmic}
\usepackage{graphicx}
\usepackage{textcomp}
\usepackage[linesnumbered,ruled]{algorithm2e}
\usepackage{array}

\newlength\mylen
\newcommand\myinput[1]{%
  \settowidth\mylen{\KwIn{}}%
  \setlength\hangindent{\mylen}%
  \hspace*{\mylen}#1\\}
  
\newcolumntype{M}[1]{>{\centering\arraybackslash}m{#1}}
  
\def\BibTeX{{\rm B\kern-.05em{\sc i\kern-.025em b}\kern-.08em
    T\kern-.1667em\lower.7ex\hbox{E}\kern-.125emX}}
\begin{document}

\title{CactusNets: Layer Applicability as a Metric for Transfer Learning\\
{}
}

\author{\IEEEauthorblockN{Edward Collier}
\IEEEauthorblockA{\textit{Dept. of Computer Science} \\
\textit{Louisiana State University}\\
Baton Rouge, Louisiana \\
ecoll28@lsu.edu}
\and
\IEEEauthorblockN{Robert DiBiano}
\IEEEauthorblockA{\textit{Ailectric}\\
Baton Rouge, Louisiana \\
robert@ailectric}
\and
\IEEEauthorblockN{Supratik Mukhopadhyay}
\IEEEauthorblockA{\textit{Dept. of Computer Science} \\
\textit{Louisiana State University}\\
Baton Rouge, Louisiana \\
supratik@csc.lsu.edu}
}

\maketitle

\begin{abstract}
Deep neural networks trained over large datasets learn features that are both generic to the whole dataset, and specific to individual classes in the dataset. Learned features tend towards generic in the lower layers and specific in the higher layers of a network. Methods like fine-tuning are made possible because of the ability for one filter to apply to multiple target classes. Much like the human brain this behavior, can also be used to cluster and separate classes. However, to the best of our knowledge there is no metric for how applicable learned features are to specific classes. In this paper we propose a definition and metric for measuring the applicability of learned features to individual classes, and use this applicability metric to estimate input applicability and produce a new method of unsupervised learning we call the CactusNet.
%
%


\end{abstract}


\section{Introduction}
Neural networks have been improving by leaps and bounds for the last decade, most notably due to the emergence of CNN, unsupervised pretraining, and better regularization methods. On many difficult image recognition tasks, they are competitive with humans \cite{delving}. Nevertheless, there is lot of room for improvement. A human can easily learn to recognize a new type of animal from just one image of the animal, or even from a crude sketch. Even if the animal is oriented differently, it can most likely be learned in one shot. The same applies to non-image data.

Humans achieve this powerful zero shot or one shot understanding via learning by analogy. In other words, they start by trying to transfer detailed knowledge from another problem, then adjusting it to "fit" the new problem where necessary. We believe learning by analogy is the most advanced form of transfer learning, and is key to achieving truly impressive results on the machine learning algorithms of the future, both on images and raw data. A human can look at a problem and consider different solutions to past problems, and intuit whether they can be applied to part or all of the new problem. Note that the applicable solutions may not belong to similar problems. For example, insight from giving a political speech might provide insight into winning a chess match; or the fact that an image is of a chess match might make it easier to track where humans are in the image. This knowledge transfer in humans is enabled by the ability to discern what previous knowledge might apply to a new problem, even with little or no labeled data.

Deep convolutional neural networks trained over large datasets learn features that are both generic to the whole dataset, and specific to individual classes in the dataset. Learned features tend towards generic in the lower layers and specific in the higher layers of a network. Methods like fine-tuning are made possible because of the ability for one filter to apply to multiple target classes. Much like the human brain, this behavior can also be used to cluster and separate classes. However, to the best of our knowledge there is no metric for how applicable learned features are to specific classes. 

We seek to measure how applicable a given network is to a given piece of data. For a convolutional network, it stands to reason that the pattern and intensity of the high level map responses should be noticeably different depending on whether many high level objects are recognized, and that how much of the image it successfully interprets is closely related to the applicability. The same principle applies to  Deep Belief Networks (DBN); it should be possible by observing the neuron responses to determine whether the network (or part of the network) is recognizing familiar patterns. By measuring applicability we can tell what problems (networks) an unknown piece of data is applicable to, even without a label. Inversely, we can immediately tell when our current network is not adequately understanding a new piece of data, and can expand, retrain, or transfer knowledge into the network immediately in real time. This ability to measure network applicability in real time will be one of the key components in learning by analogy, particularly in non-convolution networks where knowledge is abstract mathematical relationships and any knowledge may potentially apply at any layer.

\subsection{Contributions}
We aim to achieve three goals in this paper. In \cite{yosinski2014transferable}, the authors  showed that a neural network has different applicability to different problems (where a problem is a  classifying a collection of related classes). We call this notion set applicability. First, we extend this idea, defining and calculating what is to our knowledge the first metric measuring "class applicability" of a given network/layer to a single class. Extending the notions of set applicability and class applicability, it seems plausible that there is an individual image applicability for each image (or each input vector in the general case). Second, we train a second neural network to estimate this image applicability from the map responses of a convolutional neural network. We show that our method can predict high or low applicability accurately for classes and images neither of the networks has ever seen before. Finally, We demonstrate an application of this applicability measure to facilitate unsupervised learning on a   special type of pretrained deep neural network that we call the CactusNet. The CactusNet allows branching to multiple different higher level layers after each lower level layer, and uses applicability to only route input through branches that are applicable to the current input. Different output layers represent different problem types; one input can be applicable to multiple problems. When an input is not applicable to any existing problem, we create a new branch from the most applicable existing features, and start learning the remaining unknown features for the newly created problem type. 


The paper is organized as follows. In Section \ref{rw}, we briefly present the related work. The applicability metric and the CactusNet architecture will be covered in  \ref{pm}. In Section \ref{exp} we will give experimental results for both applicability and the CactusNet and our conclusions presented in Section \ref{con}. 


\section{Related Work}\label{rw}

For the past several years, advances in deep neural networks have shown to be a powerful tool for a variety of machine learning problems in multiple domains, including computer vision \cite{nam2016learning,ciregan2012multi,gidaris2015object,karki2017core}, speech \cite{graves2013speech, hinton2012deep}, and text \cite{bahdanau2014neural, kim2014convolutional}. For many of these domains, and especially for vision, each layer of the deep neural network learns features relevant to the target objective \cite{basu2017learning,basu2018deep}. For many objectives, a deep neural network requires a large-scale dataset to converge and obtain good accuracy \cite{basu2015deepsat}. For most tasks, however, large-scale datasets do not exist or are unobtainable. To circumvent this issue, existing deep neural networks can be fine-tuned for specific objectives. Fine-tuning repurposes the learned features  of a pretrained deep neural networks which then can learn the unknown features needed for the new objective. Deep convolutional neural networks (CNN) trained on ImageNet \cite{krizhevsky2012imagenet} are commonly fine-tuned for different computer vision tasks. Fine-tuning significantly reduces the amount of training examples required to converge to a target objective \cite{hinton2006reducing}.

Feature learning in deep neural networks exhibits a fascinating behavior in which the learned features tend to progress from generic, such as Gabor filters, to specific as the input moves down the network \cite{yosinski2014transferable}. Such behavior is useful in understanding how a set of features in a deep neural network can be applied to multiple objectives. This is commonly referred to as transferability or transfer learning \cite{pan2010survey}. 

Over years, researchers have worked to improve the transferability in neural networks. Deep Adaptive Networks (DAN)\cite{long2015learning} increase the transferability of task-specific features by matching different feature distributions in a reproducing kernel Hilbert space. Similar to our proposed method, DANs assume that the target dataset has little to no labeled data. DANs use multiple kernel maximum mean discrepancies (MK-MMD) \cite{gretton2012optimal} to minimize the error between two datasets to facilitate greater transferability. Our method instead quantifies how well a neural network knows or can recognize an input to facilitate unsupervised learning.

Transfer learning has also been explored for unsupervised learning as well. In survey of how transferability can be applied to unsupervised learning \cite{bengio2012deep}, the author stated that while the results look promising, transfer learning applications would improve significantly if the underlying variation in high-level features could be disentangled and made more invariant. In this work, we use applicability to demonstrate where in a network the features of an input go from invariant to variant. This point of inflection is where the CactusNet creates a branch and circumvents invariance at the more varying and more specific layers.

The human mind identifies and clusters objects based on their features regardless of whether an object is known or not \cite{grossberg2013adaptive}. Adaptive resonance theory (ART) \cite{carpenter2017adaptive,grossberg1987competitive} is a machine learning theory that attempts to determine whether an object belongs to a known object class by comparing the detected features of the object with the expected features of all known classes individually. If the smallest difference between the detected features of the object and some known class's expected features is within a set threshold then the object is classified and is considered to belong to that class. This threshold is known as the vigilance parameter. If the difference exceeds the vigilance parameter, however, the object is considered to belong to a new class. This allows ART to perform unsupervised learning as it classifies not based on a target class, but differences in features. Over the years, several new variations of ART have been proposed including Fuzzy ART \cite{carpenter1991fuzzy} which, uses fuzzy logic to improve ART's stability.

\section{Proposed Method}\label{pm}
In this section, we present the overview of our proposed method, then we outline the logic and formulation of our applicability metric in Section \ref{ap}. Next we detail how our system predicts the applicability for new or unknown classes in Section \ref{pred}. Lastly, we detail the architecture of the unsupervised CactusNet in Section \ref{cn}.

\subsection{Overview}\label{ov}

\begin{figure*}[t!]
\centering
\includegraphics[width=15cm, height=9cm]{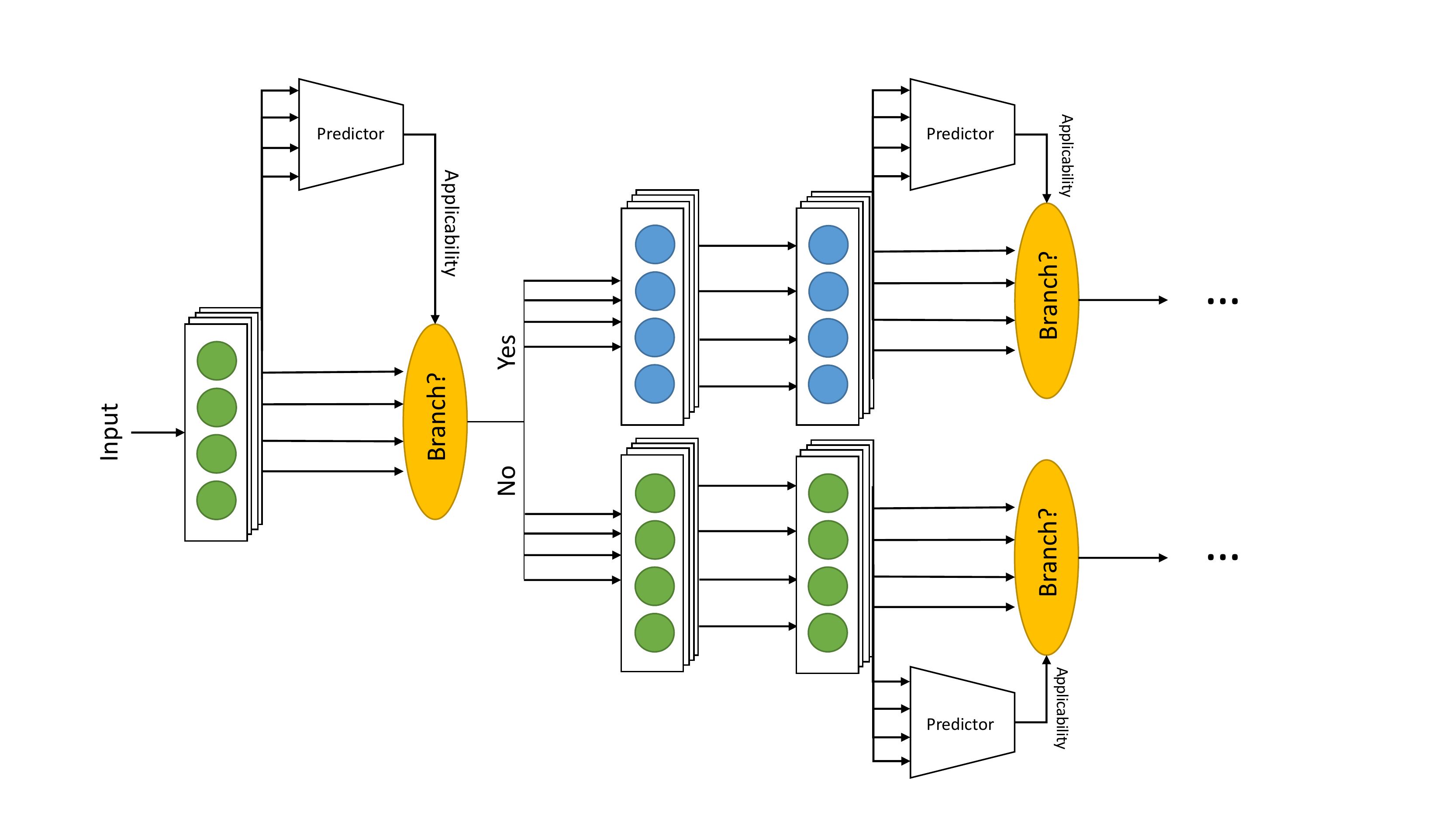}
\caption{The CactusNet architecture created in this paper. For each layer the output is sent to the applicability predictor. If the applicability is below the branch threshold then a branch is created. Otherwise the input continues down the main branch.}
\label{fig:cact1}
\end{figure*}

In our framework we seek to use our definition of applicability to facilitate unsupervised learning in deep neural networks.  We present the overall CactusNet architecture in Fig. \ref{fig:cact1}. As we see in the figure, we first build an estimator for applicability over a pretrained network. The estimator will define how applicable features learned at a particular layer are to a given input. The CactusNet will then use the applicability metric to facilitate branching for new classes.

We use the AlexNet pretrained on the ILSVRC2012 dataset \cite{krizhevsky2012imagenet} to measure applicability and create the CactusNet. The principles used in AlexNet pretrained on ImageNet can be expanded and applied to other deep neural networks, datasets, and objective. For the purposes of creating a large enough dataset of known and unknown classes, we split ImageNet in half into man-made and natural sets like in \cite{yosinski2014transferable}. The deep neural network we use is trained on the natural half, making the man-made half the unknown portion. Several natural classes were also held out from the training set to represent a group of unknown natural classes for applicability measurement. 

\subsection{Applicability}\label{ap}




We define class applicability for a trained layer in a deep neural network on an input as how well the known features can be used to differentiate the input class from all other input classes. Within an objective (classifying natural images) we identify three subsets, nonobjective unknown, objective unknown, and objective known. In the case of natural images objective known are images of classes the network has been trained on, while objective unknown are natural images of classes the network has never been trained on. Finally, nonobjective unknown are images of man made classes the network has never been trained on.

Together these three groups cover a wide range of applicabilities from low to high, allowing us to train a predictor. For a single class in this set we compare it in a series of one vs. one classifications to a separate group of classes that approximate the set of all possible inputs in a domain (be they images, sound, text, etc.)

Given a deep neural network $N$, and a number of unknown objective and nonobjective classes, we approximate the set of all possible classes, giving an unknown set, $un$, of $k=20$ classes. To find the class applicability at layer $n_i$ in a neural network $N$=\{$n_0$, $n_1$, $n_2$, ..., $n_z$\} for a given class, $x$, we measure the ability of $N$ to differentiate $x$ from all the classes in $un$. We pair $x$ with every class $un_j$ in $un$=\{$un_0$, $un_1$, $un_2$, ..., $un_{k}$\}. For each ($x$, $un_j$) pair we fine-tune $N$ with all its layers up to $n_i$ frozen, and record the test accuracy  $\xi_j$. This accuracy represents how well $x$ can be differentiated from $un_j$ using learned features from all the layers at and before $n_i$ which we will identify by the function shown $N((x,un_j), n_i)$  in  equation \ref{fun1} where $N((x,un_j), n_i)$ indicates the neural network $N$ fine-tuned with the layers $n_1, \ldots, n_i$  frozen. 

\begin{equation}
\xi_j = N((x,un_j), n_i) \label{fun1}
\end{equation}

To obtain the class applicability of $n_i$ on $x$  the function $N((x,un_k), n_i)$ must be applied to all the classes $un_k$  in $un$. Our class applicability metric is then the average differentiability between $x$ and all other $un_j$ individually. This is shown in  equation \ref{av}.

\begin{equation}
App_x = \frac{\sum_{j = 1}^{z}\xi_j}z\label{av}
\end{equation}

While  it seems plausible that there is an individual image applicability for each image,  we have not found a way to explicitly define it or measure it directly. Therefore, we set up our class applicability such that the average class applicability across all classes should approach the average separability between single pairs of classes. By extension, individual image applicability across a class should average to the class applicability. We then train a second neural network to estimate image applicability from the map responses, using class applicabilities as our labels.

\subsection{Applicability Predictor}\label{pred}

The key factor for the CactusNet architecture shown in Fig. \ref{fig:cact1} is its ability to branch at every layer for new classes. This branching is  what allows for maximum reuse of already learned features. To branch at each layer some threshold $\theta$ must be defined for each layer. We define three thresholds each corresponding to the three subsets identified for an objective (objective known, objective unknown, and nonobjective unknown). The threshold for a subset at a given layer is the average applicability across some representative sample of that subset. 

In addition to the threshold the CactusNet needs to have the ability to calculate the applicability of a given input in real-time and without sample classes from the three subsets. To calculate this applicability, predictor networks are created for each layer. For a given input within the objective of a pretrained deep neural network, the applicability network gives the predicted applicability of that layer's features for the input. 


We train the applicability predictors on large samples from our three subsets of the objective. The input for the network are the feature activations while the targets are the applicability of that specific class. The network uses a categorical cross class entropy loss function, and a relu activation function to generate an approximate applicability for an input. 


\subsection{CactusNet}\label{cn}

\begin{figure}[htbp]
\centering
\includegraphics[width=7cm, height=7cm]{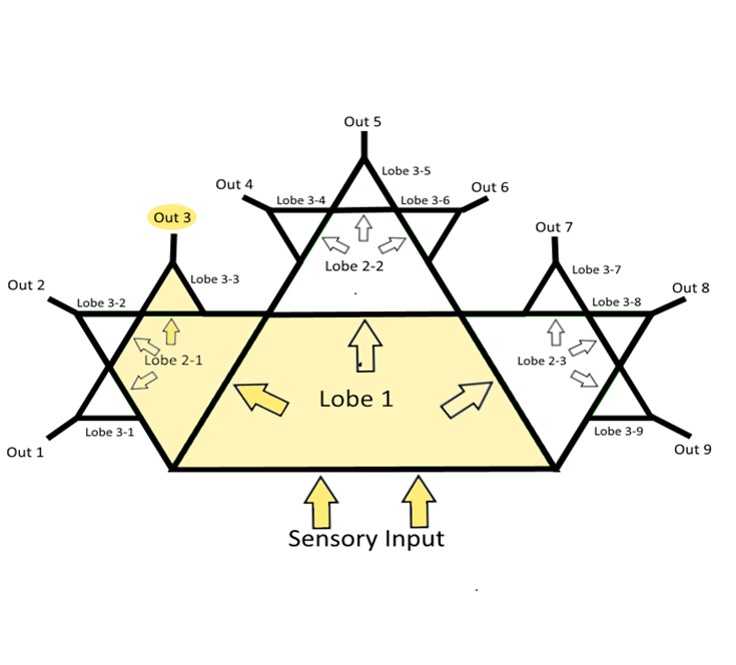}
\caption{CactusNet growth structure.}
\label{fig:cact2}
\end{figure}

The architecture of the CactusNet is shown in Figure \ref{fig:cact1}. The branching structure for the CactusNet is shown in Figure \ref{fig:cact2} and its algorithm is described in  Algorithm \ref{alg1}. The CactusNet uses the predicted image applicability of an input to determine whether the given input is either objective known, objective unknown or nonobjective unknown, and branches accordingly. The base of the CactusNet is a pretrained deep neural network. This network can be trained on any objective, and need not even be well trained for that objective. The minimum requirement is that the network has learned some features that are applicable to its target objective. The CactusNet's branching architecture is designed to efficiently handle large numbers of classes. The lower layers that contain generic features and are applicable to most every class are shared amongst all classes, minimizing the resources allocated to each new class. In the event an output layer does contain too many classes we can split by applicability and create two new branches. The applicability can be used to route inputs to the correct branch.

\begin{algorithm}
\SetKwInOut{Input}{Input}
\SetKwInOut{Output}{Output}
\caption{CactusNet Algorithm} \label{alg1}
 \Input{Network Input $x$\;}
 \myinput{List of applicability networks $A$.}
 \Output{Class label $l$.}
 \myinput{Fine-tuned network $N$.}
 initialize $l \gets \emptyset$\;
 initialize thresholds $t_1$ and $t_2$\;
 \For{i=1 to k}{
   Get features at each candidate $n_i$\;
   $app \gets n_i$ with max(applicability)\;
   \If{$app > t_1$}{
       classify normally\;
       \Return $l \gets class$}
   \Else {\If{$app \leq t_1$ and $app > t_2$}
           {$l\gets objective unknown$}}
   \Else{\If{$app \leq t_2$}
           {$l \gets$ nonobjective unknown}}
}   
\end{algorithm}


For an input $x$, we compute the feature representation for each layer $n_i$ down the network. Then we compute the predicted applicability for a layer using the feature representations and that layers applicability predictor network. If the applicability is below the threshold we have set to determine if the class is new or not, then we branch off the current network trunk at layer $n_i$.

After branching, the architecture of the network can remain the same as the original branch, or a different architecture can be used as long as it is compatible with  the shared layers in the original trunk of the CactusNet. Once a new branch is generated, then the CactusNet automatically assumes the input is of a new class and will commence learning. Given that we have not inherently modified the network architecture, the CactusNet is flexible in its operation, especially for learning. If the desire is for the CactusNet to learn from  a few training examples,  then any of the existing one shot learning methods \cite{zeroshotRomero,ZeroshotNg,Oneshot} can be used to train a new branch. If a simpler method is desired, then all classes, whether known or unknown, can be input in tiny batches for traditional learning. We chose not to focus on a particular learning method since that is a well explored topic and out of the scope of this paper.

The path an input takes down the network is determined by its applicability at each layer.  When branching occurs, the applicability threshold acts as a guide diverting inputs down the correct path. There can be any number of branches at a given layer.

Once a layer is trained, the applicability threshold of that layer is found from the applicability predictor's training results. Once a layer $n_i$ is found to be applicable, the layer output is forwarded down all branches leading off the current branch, to all candidates for layer $n_{i+1}$. Each candidate processes the data and checks its applicability to that data. If the applicability is above at least one layer's threshold, the layer with the highest applicability becomes layer $n_{i+1}$ and the input $x$ is routed there and to all branches leading off from there. If the applicability of $x$ for the layer at the end of every branch $b_i$ falls below that layer's threshold, then a new branch for $x$ will be created. The growth structure of the CactusNet is shown in Fig. \ref{fig:cact2}. In Fig \ref{fig:cact2}, we use the term lobe to denote a branch. 

\section{Experimental Evaluation}\label{exp}

In this section, we provide experimental results that illustrate the performance and functionality of our framework. Here we focus on using the ImageNet 2012 dataset with the AlexNet \cite{krizhevsky2012imagenet} architecture. We split the dataset into man-made and natural halves, and trained our deep convolutional neural network on the natural half, treating the man-made half as the unknown nonobjective portion. First we will give our experimental setup and results for applicability, showing how it compares to the transferability described in \cite{yosinski2014transferable}. We then demonstrate how image by image applicability can be predicted by convolutional applicability predictors. Lastly, we demonstrate the branching and learning features of the CactusNet.

\subsection{Datasets}
\textbf{ILSVRC2012} consists of 1.2 million images from 1000 classes. The object classes can be split between man-made and natural objects. We use the same split as described in \cite{yosinski2014transferable} that gives 449 natural classes and 551 man-made classes. The 449 natural classes were used to train a convolutional neural network to classify between them.

We defined applicability as how well a layer's features can be used to differentiate the input class from all other input classes. It would be difficult to get a representative sample of all possible input classes, so we approximate this with 20 classes our network has not been trained on, 10 natural and 10 man-made. We only use classes the network has not been trained on because any finite network would be specifically trained on a small fraction of the infinite set of all possible image classes; so a sample with all unknown classes should be more representative.

 To calculate class applicability, we separate classes into the three subsets:  unknown objective, known objective, and unknown nonobjective. In our experiments, the objective set is the natural set while the man-made represents the nonobjective set. We chose 10 classes for each set as described in \ref{ap} to train with a good mix of high, medium, and low applicability targets. The 10 classes used for the unknown objective were natural classes that  the network had not been trained on. So in all 50 classes, 30v20, were used for applicability testing:  30 classes we wanted to measure the applicability of, and 20 to approximate all possible classes to compare them against to actually find that applicability.

\subsection{Applicability Testing}

\begin{figure*}[t!]
\centering
\includegraphics[width=14cm, height=7cm]{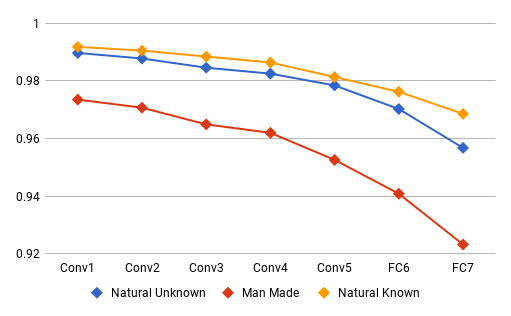}
\caption{Layer by layer applicability for the three subsets tested in our experiment. Each mark represents the average applicability over each of the 10 classes tested in the subset. The x-axis represents the bottom most layer that was frozen, whose applicability was being tested.}
\label{fig:ap}
\end{figure*}

\begin {table*}[htbp]
\caption {Layer 5 Separability}
\begin{center}
\begin{tabular}{ |p{2.8cm}||M{0.8cm}|M{0.8cm}|M{0.8cm}|M{0.8cm}|M{0.8cm}|M{1.0cm}|M{0.8cm}|M{1.0cm}|M{0.8cm}|M{0.8cm}| }
 \hline
 Class& Toilet Paper& Gong& Buckle& Bucket& Pen& Lavender& Giraffe& Mosquito& Walrus& Condor\\
 \hline
 Moth (NU) & 0.98 & 0.932 & 0.952 & 0.964 & 0.976 & 0.972 & 0.98 & 0.952 & 0.952 & 0.932\\
 Nutria (NU) & 0.972 & 0.992 & 1 & 0.98 & 1 & 1 & 0.988 & 1 & 0.948 & 0.964\\
 Screwdriver (MM) & 0.928 & 0.9 & 0.864 & 0.92 & 0.96 & 0.988 & 0.976 & 1 & 0.992 & 0.992\\
 Espresso-machine (MM) & 0.808 & 0.78 & 0.808 & 0.78 & 0.7 & 0.988 & 1 & 0.976 & 0.976 & 0.964\\
 Tabby Cat (NK) & 0.928 & 0.984 & 0.968 & 0.956 & 1 & 1 & 1 & 1 & 0.992 & 1\\
 Bee (NK) & 0.988 & 0.992 & 0.944 & 0.976 & 0.984 & 0.956 & 0.992 & 0.976 & 0.988 & 0.988\\
 \hline
 Natural Unknown &0.9708 & 0.9652 & 0.9824 & 0.9796 & 0.9888 & 0.9916 & 0.9876 & 0.9808 & 0.972 & 0.97\\
 man-made &0.8992 & 0.882 & 0.9172 & 0.9052 & 0.926 & 0.9932 & 0.986 & 0.9852 & 0.9904 & 0.992\\
 Natural known &0.9632 & 0.974 & 0.9748 & 0.9792 & 0.9816 & 0.9872 & 0.9924 & 0.9824 & 0.9852 & 0.988\\
 \hline
\end{tabular}
\bigskip
\\Table I. Separability at conv5 for several sample classes used in this paper. The set each class belongs too is marked next to the name. Each row belongs to the set of 30 classes that covered the all three objective sets, while columns are classes from the set of 20 that left out the nature known set.
\label{tab:sep}
\end{center}
\end{table*}

The applicability of a class at a specific layer is defined as the average differentiability between that class and all other classes in the unknown set. This involved training 1v1 convolutional neural networks with all the layers at, and before, the testing layer frozen. The final validation accuracy was used as the differentiability metric between the two classes. In all 600 1v1 convolutional neural networks were trained. Table \ref{tab:sep} gives an example of the layer 5 separability between six classes, one from each subset, and 10 of the test classes, while Table \ref{tab:cap} gives the applicability for 3 classes at each layer. 
%
%

\begin {table*}[htbp]
\caption {Layer by Layer Class Applicability}
\begin{center}
\begin{tabular}{ |p{2.5cm}||M{0.8cm}|M{0.8cm}|M{0.8cm}|M{0.8cm}|M{0.8cm}|M{0.8cm}|M{0.8cm}| }
 \hline
 Class& Conv1& Conv2& Conv3& Conv4& Conv5& FC6& FC7\\
 \hline
 Dolphin (NU) & 0.9876 & 0.9826 & 0.9798 & 0.9782 & 0.9726 & 0.964 & 0.953\\
 Coffee Mug (MM) & 0.9456 & 0.9444 & 0.9412 & 0.9308 & 0.9146 & 0.8998 & 0.884\\
 Llama (NK) & 0.9854 & 0.9826 & 0.9804 & 0.9782 & 0.9672 & 0.9572 & 0.9428\\
 \hline
\end{tabular}
\bigskip
\\Table II. Layer by layer class applicability for three sample classes used in this paper.
\label{tab:cap}
\end{center}
\end{table*}

In Fig. \ref{fig:ap}, we plot the average applicability for each of the three subsets as they move through the network. In the graph, we can see that the lower layers are more applicable to all the subsets, but the groups begin to separate farther along the graph. This result reinforces the results from \cite{yosinski2014transferable} where learned features start generic but become more specific the farther along the network. The features are less applicable to unknown nonobjective classes at the higher layers which would indicate the need for branching. The features are fairly applicable to unknown objective classes even at the higher layers which makes sense given that even high level features are bound to have some overlap for all classes, known or unknown, in the same objective. There is an unexpected gap between the applicability for man-made objects and the known objects at conv1. This suggests that perhaps some edge detectors are more applicable to natural images and vice versa.
%
%

\begin{figure}[htbp]
    \centering
    \begin{tabular}{cc}
  \includegraphics[width=40mm]{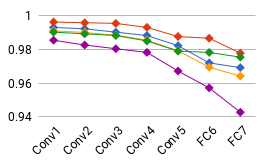} &  \includegraphics[width=40mm]{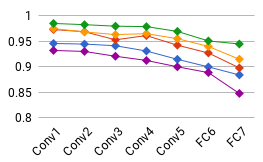} \\[6pt]
 \includegraphics[width=40mm]{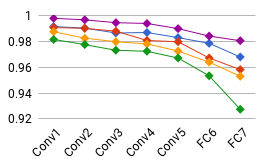} 
\end{tabular}
\caption{Layer by layer applicability for sample test classes. \textit{Top Left}: Applicability for the Natural Known set. This set tends to be more compact as the network should be applicable to all the classes. \textit{Top Right}: Applicability for the Man-Made set. We can see that the applicabilities cover a wider range than for the other classes. This indicates that the natural features are very applicable to some man-made objects. \textit{Bottom}: Applicability for the Natural Unknown set. This set is more spread out compared to its known counterpart but also more compact than the man-made set.}
\label{fig:ind}
\end{figure}

Fig. \ref{fig:ap} is broken down into smaller subsets in Fig. \ref{fig:ind}. We can further see that the spread of the applicability between classes increases between the natural and man-made sets. The average spread between the highest point and lowest point across all layers for the man-made set is 0.082, while the spread for the natural known and unknown are 0.296 and 0.21 respectively. Due to the large nature of even the natural half of ILSVRC12 there are many features that are likely applicable to even man-made objects, but there is also a gap of unknown features that would be applicable to the man-made set.

\subsection{Applicability Predictor Evaluation}

\begin{figure}[htbp]
\centering
\includegraphics[width=8cm]{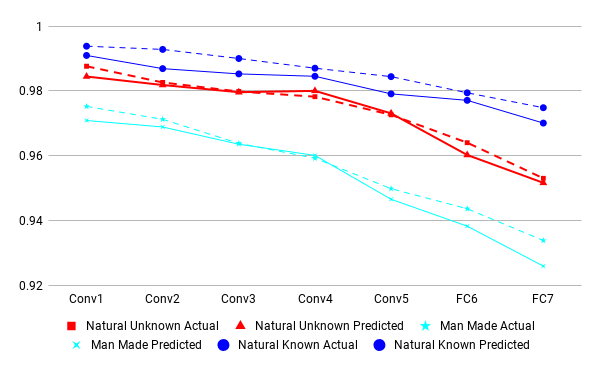}
\caption{Test results for the applicability predictor. The solid lines are the the average predicted image applicability across all test images for each class. The dashed lines are the actual class applicability computed in the 30v20 testing.}
\label{fig:data}
\end{figure}

To predict the image applicability, we used small CNN's as our applicability networks. The input for each predictor is the corresponding layer's output which is treated as a ($h$, $w$, $maps$) image where $maps$ is the number of feature maps within that layer. Each convolution block in the predictor consists of two convolutional layers and a 2x2 max pooling layer. Each convolutional layer in the first block contains 32 filters, while those in the second block contain 64. For fully connected layers we made modifications to the predictors and treated the layer output as an image of shape (1, 1, $outputs$). We train our networks by minimizing the loss function in eq. \ref{loss} where $x^i$ is the target value and $\hat{x}^i$ is the predicted value.

\begin{equation}
{\cal L} = \frac{1}{n} \sum_{i = 1}^{n}(x^i - \hat{x}^i)^2 \label{loss}
\end{equation}

The average training MSE was recorded to be 0.1485 across all the applicability predictor networks. To test the applicability predictors each predictor was tested on a class from each objective subset that the predictor was not trained on. For the test data, the recorded MSE was 0.4889. We report the training and testing mean squared (MSE) error for each layer in table. \ref{tab:mse}. 

\begin {table}[htbp]
\caption {Test vs Validation MSE}
\begin{center}
\begin{tabular}{ |p{1.0cm}|p{1.0cm}|p{1.2cm}| }
 \hline
 Test& Layer& Validation\\
 \hline
 0.1893 & Conv1 & 0.3108\\
 0.1538 & Conv2 & 0.6137\\
 0.1238 & Conv3 & 0.6632\\
 0.1657 & Conv4 & 0.4252\\
 0.1433 & Conv5 & 0.4857\\
 0.1253 & FC6 & 0.3883\\
 0.1382 & FC7 & 0.5344\\
 \hline
\end{tabular}
\bigskip
\\Table III. Image applicability network mean squared error for both the testing and validation datasets.  
\label{tab:mse}
\end{center}
\end{table}

Fig. \ref{fig:data} shows the layer by layer predicted applicability versus the actual applicability metric we computed for that class. We can see in the graph that our applicability predictors are able to produce an image applicability that is very close to its true value. This suggests that our applicability networks are able to distinguish if the CNN has or has not been trained on an image even if they are both highly applicable. In Table \ref{tab:res}, we give sample results from the conv4 applicability predictor. Interestingly the predictions on the man-made dataset tend to have the most variance, but none of the predictions overshoot the target value by very much.  

\begin{figure}[htbp]
\centering
\includegraphics[width=9cm]{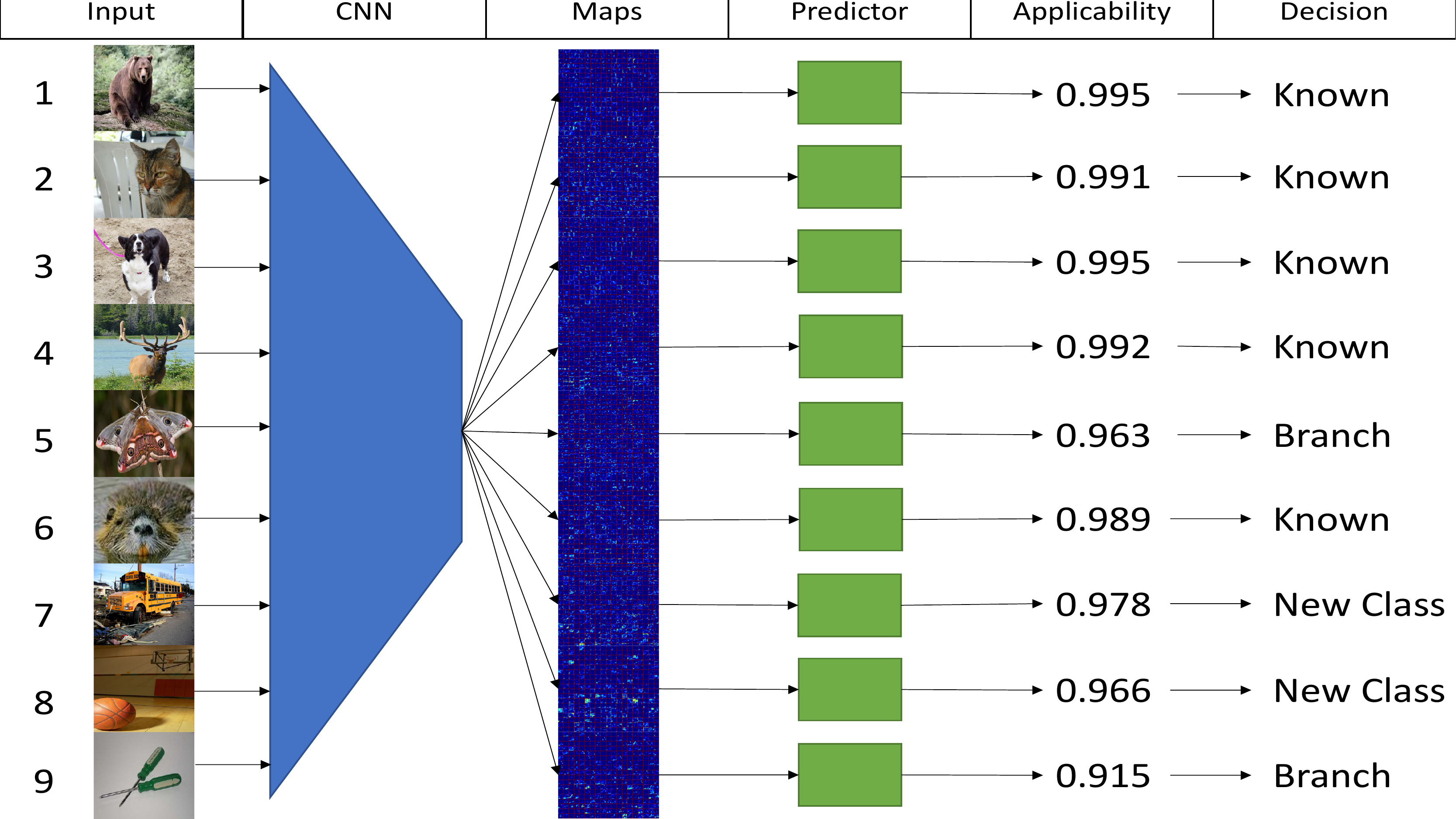}
\caption{Work flow for applicability predictions in our CactusNet test. The feature maps are extracted from the CNN and fed into the applicability predictors, which then produce a predicted image applicability.}
\label{fig:samp}
\end{figure}

\begin {table}[htbp]
\caption {Test Image Applicability (\%)}
\begin{center}
\begin{tabular}{ |p{1.7cm}|p{0.8cm}|p{1.0cm}|p{0.7cm}| }
 \hline
 Objective& Actual& Predicted& Error\\
 \hline
Nat Unknown & 97.82 & 97.95 & 0.13\\
Nat Unknown & 97.82 & 97.95 & 0.13\\
Nat Unknown & 97.82 & 98.26 & 0.44\\
Nat Unknown & 97.82 & 97.90 & 0.08\\
Nat Known & 98.70 & 98.84 & 0.14\\
Nat Known & 98.70 & 97.91 & 0.79\\
Nat Known & 98.70 & 98.82 & 0.12\\
Nat Known & 98.70 & 98.44 & 0.26\\
Man-Made & 95.92 & 96.70 & 0.78\\
Man-Made & 95.92 & 97.31 & 1.36\\
Man-Made & 95.92 & 94.80 & 1.12\\
Man-Made & 95.92 & 95.34 & 0.58\\

 \hline
\end{tabular}
\bigskip
\\Table IV. Actual class applicability versus the predicted image applicability for 12 sample images from three classes each from one of the objective sets.
\label{tab:res}
\end{center}
\end{table}

\subsection{Convolutional CactusNet}
With trained applicability predictor networks for each layer we use the pretrained AlexNet to create a convolutional CactusNet. We define two applicability thresholds, $\tau_1$ and $\tau_2$. A higher threshold is used to separate images that are applicable but not from a known class, giving us the potential recognize and add new classes in an unsupervised manner. A lower threshold is used to recognize image that are not applicable to the current layer and facilitate branching.

\begin{equation}
\tau_n = q - \frac{1}3(q - y_n) \quad\text{where}\quad n = \{1, 2\} \label{thres}
\end{equation}

In our experiments at conv4, we used the natural known applicability at conv4, 0.986, for $q$, our upper baseline. Our lower baselines, $y_1 = 0.983$ and $y_2 = 0.923$, are taken from natural unknown at conv4 and man-made at layer fc7 respectively. This gives us $\tau_1 = 0.985$ and $\tau_2 = 0.965$.

With our branching threshold set we were able to create a convolutional CactusNet that identified classes as either known, unknown but applicable or non-applicable. The CactusNet branches and fine-tunes based on which objective set an input image exists in. Figure \ref{fig:samp} gives the work flow we used for the applicability predictors to facilitate branching at conv4 of AlexNet. In our experiments, we passed eight images of varying applicability through our CactusNet, shown in Figure \ref{fig:samp}. Images 1, 2, 3, 4 and 6 showed applicability greater than $\tau_1$ and were detected as known. These were all furry mammals, which ImageNet has been extensively trained with. Images 7 and 8 were below $\tau_1$ but above $\tau_2$, meaning our network is fairly applicable but should define new class labels for them. Images 5 and 9 (moth and screwdriver) were below $\tau_2$, and in these cases the CactusNet would branch because it is not applicable to these unusual shapes. Because branching is automatically determined with thresholding the CactusNet grows in an unsupervised manner. While branching is a possibility at every layer, branching realistically only starts to occur at and beyond conv4 where the features stop being generic. 

\section{Conclusion}\label{con}
In this paper we presented, to  the best of our knowledge,  the first ever definition and metric for the applicability of a deep neural network's layer features to a specific input. We showed that the results of our metric are in line  with the expected transferability of features in deep neural networks. We also demonstrated how small predictor neural networks can be used to predict the applicability for an input in real-time. Lastly, we used our applicability metric to create an efficient and self-growing deep neural network, called the CactusNet, that can perform unsupervised learning with efficient reuse of learned parameters.

\section{Acknowledgments}\label{ack}
We thank Richard J. Barbalace (Ailectric) for suggestions that improved the manuscript.

\bibliographystyle{IEEEtran}  
\bibliography{main.bib}

\end{document}